\documentclass[10pt,twocolumn,letterpaper]{article}

\usepackage[pagenumbers]{cvpr} 

\usepackage{graphicx}
\usepackage{amsmath}
\usepackage{amssymb}
\usepackage{booktabs}
\usepackage{xcolor}
\usepackage{bbding}
\usepackage{physics}

\usepackage[pagebackref,breaklinks,colorlinks]{hyperref}

\usepackage[capitalize]{cleveref}
\crefname{section}{Sec.}{Secs.}
\Crefname{section}{Section}{Sections}
\Crefname{table}{Table}{Tables}
\crefname{table}{Tab.}{Tabs.}

\begin{document}

\title{High Quality Segmentation for Ultra High-resolution Images}

\author{
    Tiancheng Shen$^{1}$, 
    Yuechen Zhang$^{1}$,
    Lu Qi$^{1}$,
    Jason Kuen$^{2}$,
    Xingyu Xie$^{3}$ \\ 
    Jianlong Wu$^{4}$,
    Zhe Lin$^{2}$,
    Jiaya Jia$^{1}$
    \\
    $^1$The Chinese University of Hong Kong~~~~
    $^2$Adobe Research\\
    $^3$Peking University~~~~
    $^4$Shandong University\\
}

\maketitle

\begin{abstract}

To segment 4K or 6K ultra high-resolution images needs extra computation consideration in image segmentation. Common strategies, such as down-sampling, patch cropping, and cascade model, cannot address well the balance issue between accuracy and computation cost. Motivated by the fact that humans distinguish among objects continuously from coarse to precise levels, we propose the Continuous Refinement Model~(CRM) for the ultra high-resolution segmentation refinement task. CRM continuously aligns the feature map with the refinement target and aggregates features to reconstruct these images' details. Besides, our CRM shows its significant generalization ability to fill the resolution gap between low-resolution training images and ultra high-resolution testing ones. We present quantitative performance evaluation and visualization to show that our proposed method is fast and effective on image segmentation refinement. Code will be released at \href{https://github.com/dvlab-research/Entity}{https://github.com/dvlab-research/Entity}.

\end{abstract}

\section{Introduction}
\label{sec:Introduction}

With the rapid development of camera and display equipment, the resolution of images is getting higher and higher, where 4K and 6K resolutions become common. It gives different chances in portrait photo post-processing, industrial defect detection, medical diagnose, \etc. However, ultra high-resolution images also bring challenges to the classical image segmentation methods. First, the significant number of input pixels is computationally expensive and GPU memory-hungry. Second, most existing methods up-sample the final prediction for 4 to 8 times through interpolation~\cite{yang2018denseaspp, yuan2018ocnet, zhao2017pyramid,chen2018encoder,zhao2018psanet}, without building fine-grained details on output masks.

Previous segmentation refinement methods include those of~\cite{huynh2021progressive, yuan2020segfix, lin2017refinenet, kirillov2020pointrend}. They still target at images with 1K$\sim$2K resolutions. Work of~\cite{cheng2020cascadepsp, yang2020meticulous} handles ultra high-resolution refinement based on low-resolution masks generated from classic segmentation algorithms. They utilize cascade-scheme in decoder to upsample intermediate refinement results in several resolution stages until reaching the target resolution. They are still time-consuming due to working in discrete style on predefined resolution stages of decoder. We instead consider continuity to make the decoding more efficient and more friendly to the learning of up-sampling resolution. We propose the Continuous Refinement Model~(CRM) to exploit continuity.

\begin{figure}[t]
	\centering
	\includegraphics[width=0.95\linewidth]{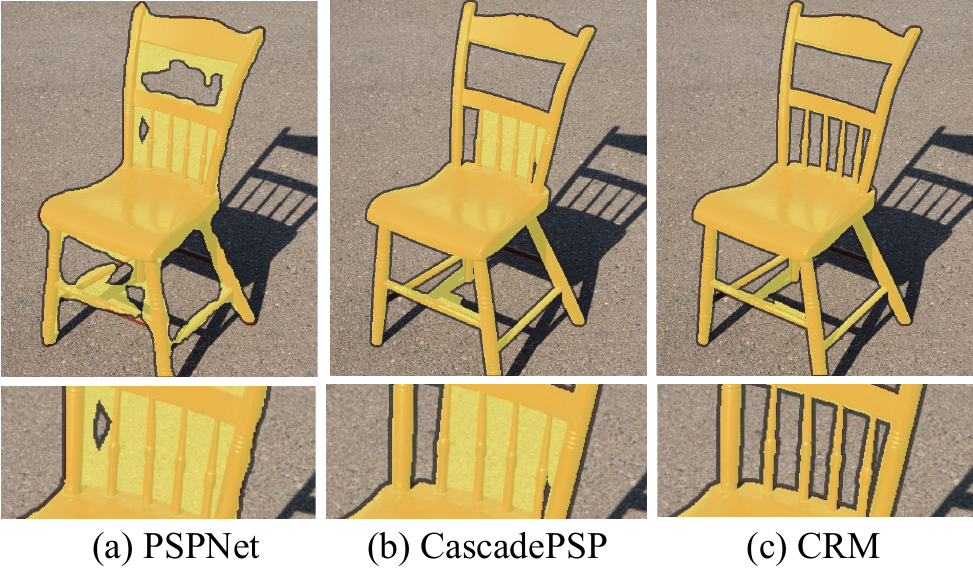}
	\caption{Coarse mask refinement results. (a) Coarse mask from PSP~\cite{zhao2017pyramid}, (b) refined mask of state-of-the-art \cite{cheng2020cascadepsp}, and (c) refined mask of our proposed CRM. The image is from BIG~(2K$\sim$6K res).
	}
	\label{fig:teaser_image}
\end{figure}

The coarse mask is from low-resolution segmentation. In order to expand it further, the problem is similar to a classical super-resolution~(SR) task. Other than classical SR methods, constructing continuous local representation is proposed~\cite{chen2021learning}. We note that utilizing implicit function~\cite{mildenhall2020nerf} to handle high-resolution segmentation refinement is not trivial. First, the resolution of the training image in our task is around 500, while the training image for SR is with 2K resolution. The training strategy to down-sample the input to SR would make our input mask tiny and meaningless. Second, more multi-level semantic features are needed compared with super-resolution configuration. Third, there exists a resolution gap between training on low-resolution and testing on ultra high-resolution. Therefore, this task needs specific designs. %

To realize the continuity in ultra high-resolution segmentation refinement, we first propose Continuous Alignment Module~(CAM) to align the feature and refinement target continuously (different from utilizing the cascade scheme in decoder). In CAM, the coordinates of feature and refinement target are transferred into a continuous space. We then align position and feature based on the continuous coordinate. An implicit function combines position information and aligned latent image feature to predict the segmentation label for the queried pixel on images. Here, the pixel-wise implicit function models the relationship between continuous position and prediction and realizes image-aware refinement by latent feature.  Overall, this design is simpler and lighter than the cascade-based decoder, but generates more precise refinement mask as \cref{fig:teaser_image}. 

In addition, there is a resolution gap between low-resolution training images and ultra high-resolution testing ones. In cascade-decoder-based methods~\cite{cheng2020cascadepsp, yang2020meticulous}, convolution always covering a fixed size neighbor patch under the training resolution reduces its generalization to other testing resolutions. However, implicit function in CRM is in pixel-wise extracted feature without this bias. Also, in our multi-resolution inference strategy, low-resolution input is inferred first. Then we increase the input resolution to generate more details in the refined mask. Working with a multi-resolution inference strategy, CRM realizes stronger generalization ability than previous methods~\cite{cheng2020cascadepsp}. In experiments, our CRM achieves better performance and infers more than twice as fast as previous state-of-the-art methods in the ultra high-resolution segmentation refinement task. 

\begin{figure}[t]
  \centering
  \includegraphics[width=0.98\linewidth]{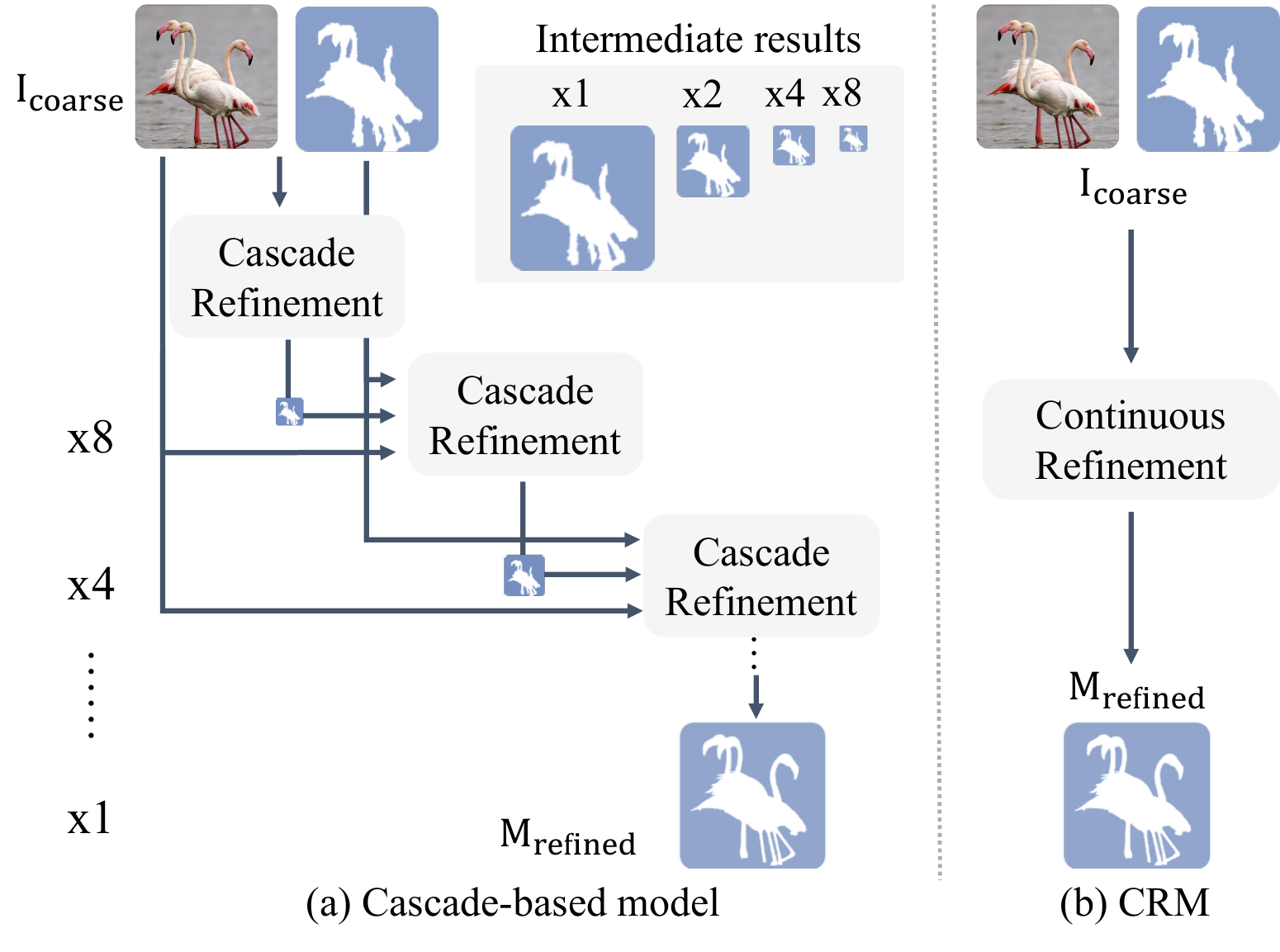}
  \hfill
  \caption{Structure difference between (a) Cascade-based decoder in model~\cite{cheng2020cascadepsp} and (b) our CRM. We can see CRM is much simpler, which is the base of our speed advantage.}
  \label{fig:CascadePSPandCRM}
\end{figure}

Our main contribution is the following.
\begin{itemize}
\item
We propose a general Continuous Refinement Model~(CRM). It introduces an implicit function that utilizes continuous position information and continuously aligns latent image feature in ultra high-resolution segmentation refinement. Without a cascade-based decoder, we effectively reduce computation cost and yet reconstruct more details.
\item
CRM with multi-resolution inference is suitable for using low-resolution training images and ultra high-resolution testing images. Due to the simple design, even with refining from low- to high-resolution, the total inference time is less than half of  CascadePSP~\cite{cheng2020cascadepsp}.
\item 
In experiments, CRM yields the best segmentation results on ultra high-resolution images. It also helps boost the performance of state-of-the-art panoptic segmentation models without fine-tuning.
\end{itemize}

\section{Related Work}
\subsection{Semantic Segmentation}
Semantic segmentation is to assign a class label to each pixel for an image. FCN~\cite{long2015fully} introduces the deep convolution network into semantic segmentation and achieved remarkable progress, and deep convolution networks are the dominant solution in this area. Later work includes PSPNet~\cite{zhao2017pyramid}, DeepLab series methods~\cite{chen2017deeplab, chen2017rethinking, chen2014semantic, chen2018encoder}, and other outstanding work \cite{fu2019dual, yuan2018ocnet, wang2018non, vaswani2017attention, wei2017object, yang2018denseaspp, he2019adaptive, lin2019zigzagnet, gidaris2017detect, islam2017label, li2020spatial}.

\begin{figure*}[t]
  \centering
  \includegraphics[width=0.9\linewidth]{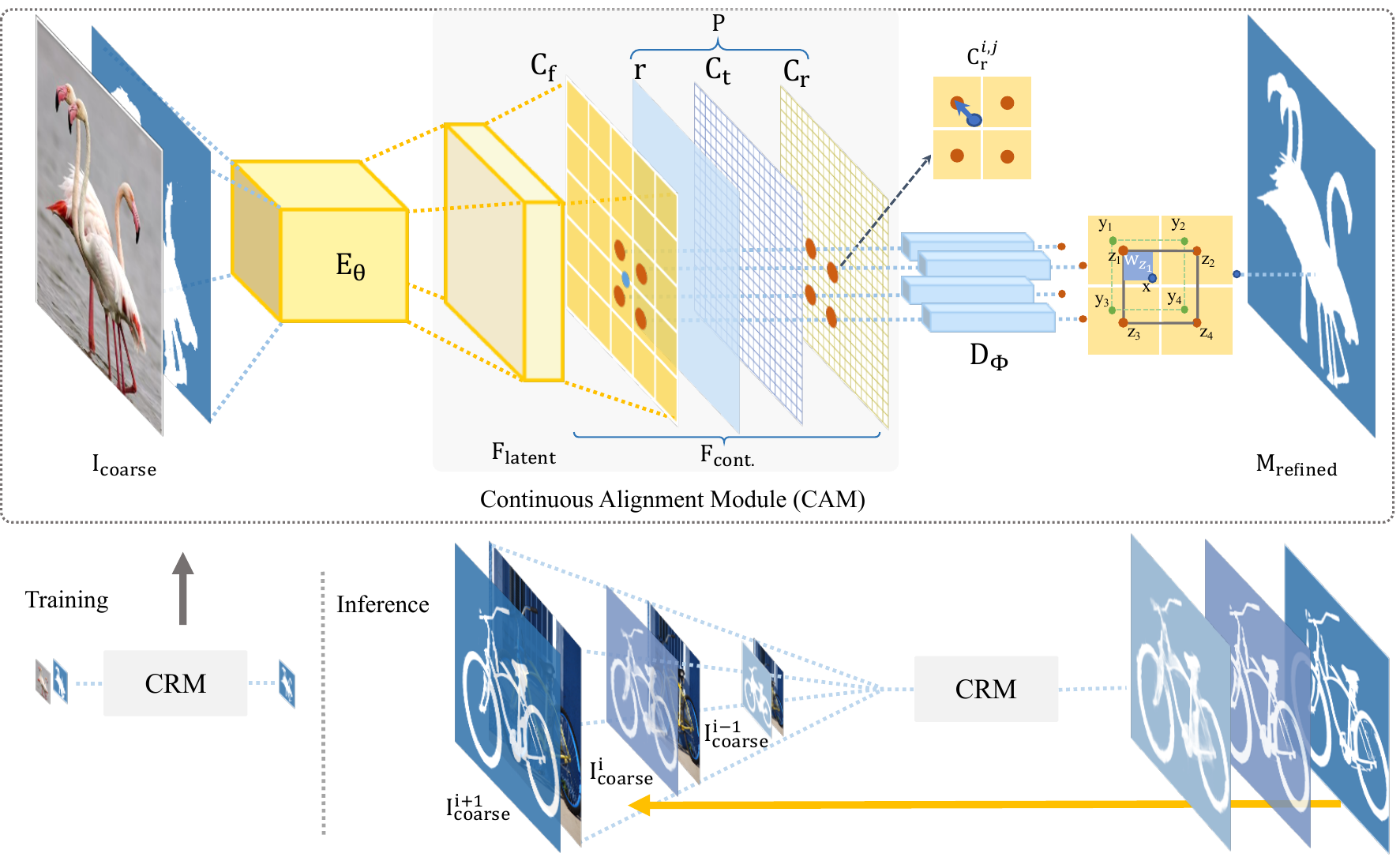}
  \hfill
  \caption{The general framework of CRM. The upper part is the structure of the model. The lower part is the training and testing process of CRM. From the lower part, we can also see the resolution gap between low-resolution training and high-resolution testing.}
  \label{fig:GeneralFramework}
\end{figure*}

Among these methods, output stride (or down-sample ratio) is one point that cannot ignore. In most semantic segmentation methods, it is set to 4$\times$~\cite{yang2018denseaspp, yuan2018ocnet} or 8$\times$\cite{zhao2017pyramid,chen2018encoder,zhao2018psanet}, which reduces precision. Directly interpolating prediction logits to target-size results in jagged edge and fewer details. In contrast, our proposed CRM continuously aligns features to arbitrary target refinement resolution, which is more natural for visual instinct and friendly to detail reconstruction.

\subsection{Segmentation Refinement}
\label{sec:SegmentationRefinement}
The segmentation refinement technique is proposed to improve the quality of image segmentation. In this track, recent work can be categorized into two classes according to the image size of high-resolution~(1K$\sim$2K) or ultra high-resolution~(4K$\sim$6K). 

For the refinement techniques of images around 1K resolution, they greatly improve the segmentation quality. The remaining drawbacks include graphical models adhering to low-level color boundaries~\cite{chen2014semantic, zheng2015conditional}, propagation-based approaches facing computational and memory constraints~\cite{liu2017learning}, and large models prone to overfitting while shallow refinement networks with limited refinement capability~\cite{lin2017refinenet, kirillov2020pointrend, huynh2021progressive}.

This paper focuses on ultra high-resolution image segmentation refinement on, e.g., 4K images. Due to this resolution setting, the above methods would face resource and effectiveness difficulties. Cascade-in-decoder methods~\cite{cheng2020cascadepsp, chen2019collaborative} achieve the state-of-the-art refinement performance on ultra high-resolution images due to its cascade network structure~\cite{he2019bi, sun2013deep, zhao2018icnet, shen2019r} and a global-local patch-based refining pipeline. 

However, the heavy cascade structure in the decoder needs down-sampling and cropping patches during inference, which increases cost, loses details, and destroys global context. To solve these problems in ultra high-resolution image segmentation, we propose CRM. Through CAM in CRM, we continuously align the feature map with refinement target simply and elegantly.  
The structure difference between cascade-based model~\cite{cheng2020cascadepsp} and our CRM is presented in \cref{fig:CascadePSPandCRM}.

\subsection{Implicit Function for Representation}
\label{sec:relatedImplicitFunction}
In the beginning, implicit neural is designed to represent an object or a scene in a neural network~(by usually multi-layer perceptron), which maps coordinates to different signals. For example, NeRF~\cite{mildenhall2020nerf} maps the 3D coordinate and 2D view angle into RGB and transparency of certain positions from specific views. PixelNerf~\cite{yu2021pixelnerf} introduces an architecture that conditions a NeRF~\cite{mildenhall2020nerf} on image input in a fully convolutional manner, which realizes scene-aware modeling. In addition, its  ``relative camera poses" idea also inspires research to use relative position information. 

As another extension, Semantic-NeRF~\cite{zhi2021place} extends neural radiance fields to encode semantics with appearance and geometry jointly. The intrinsic multi-view consistency and implicit function's smoothness benefit segmentation by enabling efficient propagation on sparse and noisy labels. There is work utilizing implicit functions in 2D image~\cite{sitzmann2020implicit, dupont2021coin, chen2021learning,shaham2021spatially, chen2019learning}. To the best of our knowledge, we make the first attempt to introduce implicit functions to segmentation. 

\section{Proposed Method}
This section first describes the general framework for the Continuous Refinement Model~(CRM), then illustrates the Continuous Alignment Module~(CAM) and the following implicit function. Finally, we introduce the corresponding inference strategies to exploit continuity in ultra high-resolution.

\subsection{General Framework}
\label{sec:framework}
As illustrated in \cref{fig:GeneralFramework}, following the setting of CascadePSP~\cite{cheng2020cascadepsp}, our proposed CRM takes an image $I \in \mathbb{R}^{3\times H\times W}$ and a coarse segmentation mask $M_{\text{coarse}} \in \mathbb{R}^{1\times H\times W}$ as input. First, $I$ and $M_{\text{coarse}}$ are concatenated as $I_{\text{coarse}} \in \mathbb{R}^{4\times H\times W}$ and are represented as latent embedding $F_{\text{latent}} \in \mathbb{R}^{C\times h\times w}$ by an encoder $E_{\theta}$ as \cref{eq:cam0}, where $\theta$ denotes the parameters. 
\begin{equation}
F_{\text{latent}}=E_{\theta}(I_{\text{coarse}}).
\label{eq:cam0}
\end{equation}

Second, $F_{\text{latent}}$ and position information $P$ are continuously aligned to be the target size feature $F_{\text{cont.}} \in \mathbb{R}^{(C+6) \times H\times W}$ through CAM without explicit up-sampling as \cref{eq:cam1}, where $[\cdot,\cdot]$ denoted concatenation. 
\begin{equation}
F_{\text{cont.}}=\mathrm{CAM}([P, F_{\text{latent}}]).
\label{eq:cam1}
\end{equation}

Finally, $F_{\text{cont.}}$ passes an implicit-function-based decoder $D_{\phi}$ and feature aggregation step, making refined mask $M_{\text{refined}}$ generated as below:
\vspace{-1mm}
\begin{equation}
M_{\text{refined}}(x) = \sum_{\text{z}_{\text{k}} \in N(x)} \frac{\text{w}_{\text{z}_{\text{k}}}}{\sum \text{w}_{\text{z}_{\text{k}}}} D_{\phi}(F_{\text{cont.}}(\text{z}_{\text{k}})),
\label{eq:cam2}
\end{equation}
where $x$ is an aligned point, $N(x)$ denotes the set of $x$'s supporting points $\text{z}_{\text{k}}, k \in \{1, 2, 3, 4\}$, $w_{z_{k}}$ is the aggregation weights~(
swap the area value of the box between $x$ and $\text{z}_{\text{k}} \in N(x)$ symmetrically with x as the center), and $F_{\text{cont.}}(\text{z}_{\text{k}})$ is the feature vector of $\text{z}_{\text{k}}$ on $F_{\text{cont.}}$.

\subsection{Continuous Alignment Module}
\paragraph{Motivation}
After passing the image encoder, the size of the encoded feature is smaller than the refinement target. Intermediate feature or refined results need to be up-sampled to later stages progressively. In previous work~\cite{cheng2020cascadepsp, yang2020meticulous} on ultra high-resolution image segmentation, cascade scheme seems an indispensable part of the decoder.  
Although novel designs alleviate information damage after up-sampling in a specific resolution, the overall process is hard to restore more details. 

We note that the discrete manner in cascade-based decoder with predefined up-sampling ratios can be regarded as constraints to up-sampling, limiting the further improvement and reducing generality. In addition, it increases the complexity of the whole framework, illustrated in \cref{fig:CascadePSPandCRM}. Our proposed Continuous Alignment Module~(CAM) utilizes position information and feature alignment to model the continuous deep feature $F_{\text{cont.}}$.

\vspace{-1mm}
\paragraph{Position Information~$P$}
Referring to NeRF-Series~\cite{mildenhall2020nerf, yu2021pixelnerf, zhi2021place}, the position information is the essential input to the implicit function. Coordinate of refinement target $C_{\text{t}}$ is projected to feature map coordinate $C_{\text{f}}$. This operation creates continuous coordinates for pixels on different resolution feature maps and various desired inference resolutions, shown in \cref{sec:Inference}. 

The absolute coordinate may vary with the image and feature size. To make our CRM universal for images of arbitrary sizes, the $C_{\text{t}}$ and $C_{\text{f}}$ are normalized to certain range $[-1, 1]$. After projection, the offset between the points on $C_{\text{t}}$ and their corresponding nearest points on $C_{\text{f}}$ is denoted as $C_\text{r}$. In \cref{fig:GeneralFramework}, the $C_{\text{r}}^{i, j}$ represents the offset~(blue arrow) on position$(i, j)$. The relative target coordinate offset $C_\text{r}$, the ratio $r$ between feature and target ~\cite{chen2021learning}, and the refinement target position $C_{\text{t}}$ form the position information $P$ as
\vspace{-1mm}
\begin{equation}
P=\left \{C_{\text{r}}, r,  C_{\text{t}}\right \}.
\label{eq:position}
\end{equation}
The continuous position information is the basis of continuity in CRM.

\vspace{-2mm}
\paragraph{Continuous Feature Alignment} 
Compared with continuous resolution conversion in SR~\cite{chen2021learning}, $F_{\text{latent}}$ from $E_{\theta}$ in the equation needs to enhance by fusing global-local information for the segmentation refinement task. For simplicity, $F_{\text{latent}}$ includes the enhancement. The refinement target position $C_{t}$ can also be regarded as a global feature.
Then, same as that for the position information, we align each pixel in refinement target to $F_{\text{latent}}$. The continuous feature $F_{\text{cont.}}$ is established by concatenating the position information $P$ and the aligned $F_{\text{latent}}$ as shown in \cref{eq:cam1}.

Therefore, compared with discrete resolution conversion, CAM up-samples feature in a continuous manner. The discrete predefined up-sampling ratios reduce the learning difficulty but constrain the up-sampling process. Out CAM has a greater degree of freedom in this respect, which means a larger space to optimize and higher performance potential. The multi-resolution inference in~\cref{sec:Inference} gives full play to the advantage of continuity of CAM. 


\subsection{Implicit Function in CRM}
After CAM, implicit-function $D_{\phi}$ takes $F_{\text{cont.}}$ as input. The reason to utilize implicit function is its impressive ability to process continuous coordinates and reconstructing details~\cite{mildenhall2020nerf, chen2021learning, yu2021pixelnerf, zhi2021place}. 

A queried point~(blue point on ~\cref{fig:GeneralFramework}) on target refinement mask could be denoted as $x(i, j)$, in which $(i, j)$ is its unnormalized position. First, we find its neighbor points $y_{k}, k \in \{1, 2, 3, 4\}$~(green points on ~\cref{fig:GeneralFramework}) on target refinement mask, whose positions are $(i\pm1, j\pm1)$. Next, the nearest points of $y_{k}$, denoted as $z_k$~(red points on ~\cref{fig:GeneralFramework}), are selected on the aligned feature map. And $z_{k}$ are utilized as the supporting points of $x$, represented as $N(x)$.
We then input $z_{k}$'s feature vector $F_{\text{cont.}}(z_{k})$ to implicit function $D_{\phi}$. Finally, we aggregate the implicit function's output. The aggregation weights, i.e., area value~$w_{z_{k}}$, are calculated from relative coordinate offsets $C_{r}$ in \cref{eq:cam2}. The aggregated output is the final prediction result on $(i, j)$.

\vspace{-2mm}
\paragraph{Analysis} It is well-known that the forward process of CNNs~(e.g., CascadePSP~\cite{cheng2020cascadepsp}) and MLPs~(e.g., CRM) can be regarded as a sequence of operations built on matrix-vector multiplications and nonlinear activation. At initialization, all the weights are sampled from well-scaled Gaussian. Hence, each layers' feature shares almost the same Eculiclid norm with high probability (see Cor. A.10 in ~\cite{allen2018convergence}). Namely, for some constant $c$, with probability at least $1-2 \exp \left(-c\varepsilon^{2} m\right)$, we have:
\begin{equation}
\|\phi(A F_{\text{cont.}})\|_{2} \in \qty(1\pm\varepsilon)\qty(\|F_{\text{cont.}}\|_2),
\label{eq:norm}
\end{equation}
where each entry of the matrix $A \in \mathbb{R}^{d\times m}$ is sampled from $\mathcal{N}(0,\frac{1}{m})$, $F_{\text{cont.}}$ is the fixed feature~(same as $F_{\text{cont.}}$ in \cref{eq:cam1}), $\varepsilon \in [0,1]$, $\|\cdot\|_2$ is $\ell_2$-norm, and $\phi: \mathbb{R} \to \mathbb{R}$ is the ReLU activation.
\par
The norm is almost preserved after going through one layer. However, if we further append one operation of weighted average on $\phi(A F_{\text{cont.}})$, things become interesting. 
The appending weighted average can always help to improve the representation ability of model, i.e., 
\begin{equation}
\operatorname{dim}(\sum_{\text{z}_{\text{k}} \in N(x)} \frac{\text{w}_{\text{z}_{\text{k}}}}{\sum \text{w}_{\text{z}_{\text{k}}}} \phi(A F_{\text{cont.}}(\text{z}_{\text{k})})) \geq \operatorname{dim}(\phi(A F_{\text{cont.}})),
\label{eq:dim}
\end{equation}
where $\operatorname{dim}$ is the dimension of space.

A toy example is that, when $F_{\text{cont.}}$ is the $m$-dimensional sphere $\mathcal{S}(m)$, $\phi(A F_{\text{cont.}})$ will concentrate around the sphere $\mathcal{S}(d)$ by the norm-preserving property. However, after combining with the weighted average operator, we can get any points \emph{in} the $d$-dimensional ball $\mathcal{B}(d)$.  
Generally, $\operatorname{dim}\qty(\mathcal{B}(d))>\operatorname{dim}\qty(\mathcal{S}(d))$.

Back to the section, the main difference between CRM and CascadePSP~\cite{cheng2020cascadepsp} is the decoder part. Take four neighboring points as an example. CRM utilizes MLP and area-based average instead of 2$\times$2 convolution. Therefore, the dimension of CRM's feature space is larger. If the four points all belong to the same class, the influence is not very large. Still, for boundary region, where 4 points belonging to different classes, 
larger feature space always provides more distinguishable feature to classified. 
From this view, we can give some hints about CRM having stronger boundary region representation and predicting better details.

\subsection{Training and Inference Strategy}
\paragraph{Training without Cascade}
LIIF~\cite{chen2021learning} proposes an elegant solution for SR with the implicit function. 
It has 2K images as ground truth and generates any low-resolution images as input.
Ultra high-resolution images with precise segmentation annotations are too few to train. In addition, high-resolution training is directly limited by the constraint of GPU memory and batch size. 

With these challenges, we follow the training setting of CascadePSP~\cite{cheng2020cascadepsp} to use low-resolution images in their initial resolution. $M_{\text{coarse}}$ is generated by morphological perturbation of the provided ground truth mask $M_{\text{gt}}$. 
We design the training loss in a simple way on the final prediction $M_{\text{refined}}$ without different loss functions on different resolution stages~\cite{cheng2020cascadepsp}. Our loss term $L(\theta, \phi)$ is calculated on the refinement target as
\begin{equation}
L(\theta, \phi)=\sum_{i=1}^{4}{w_{\text{i}} \cdot L_{\text{i}}\qty(M_{\text{refined}},M_{\text{gt}})},
\end{equation}
where $L_{i}, i \in [1,2,3,4]$ denote cross-entropy loss, L1 loss, L2 loss, and gradient loss, respectively. $w_{i}$ are their corresponding weights. $(\theta, \phi)$ are the parameters of encoder $E_{\theta}$ and decoder $D_{\phi}$. $M_{\text{gt}}$ denotes the ground truth mask.

Although we train on the low resolution, multi-resolution inference strategy exploits the continuity potential and narrows the training and testing resolution gap. 

\begin{table*}[!htp]
\centering
\begin{tabular}{lccccc}
\toprule
IoU/mBA &  
Coarse Mask &   
SegFix\cite{yuan2020segfix} &
MGMatting\cite{mgmatting} & 
CascadePSP\cite{cheng2020cascadepsp} &
CRM(Ours)  \\ \toprule
FCN-8s~\cite{long2015fully}    &  72.39/53.63 & 72.69/55.21 & 72.31/57.32 & 77.87/67.04 & \textbf{79.62}/\textbf{69.47} \\ 
DeepLabV3+~\cite{chen2018encoder} &  89.42/60.25 & 89.95/64.34 & 90.49/67.48 & \textbf{92.23}/74.59  & 91.84/\textbf{74.96}  \\ 
RefineNet~\cite{lin2017refinenet} &  90.20/62.03 & 90.73/65.95 & 90.98/68.40 & 92.79/74.77 & \textbf{92.89}/\textbf{75.50} \\ 
PSPNet~\cite{zhao2017pyramid} &  90.49/59.63 & 91.01/63.25 & 91.62/66.73 & 93.93/75.32 &  \textbf{94.18}/\textbf{76.09} \\ \hline
Average Improve. & 0.00/0.00   & 0.47/3.30   &       0.73/6.10      &     3.58/14.05        &        \textbf{4.01}/\textbf{15.12}                              \\  \toprule
\end{tabular}
\vspace{-2mm}
\caption{IoU and mBA results on the BIG dataset comparing with other mask refinement methods. Coarse mask is from FCN, DeepLabV3+, RefineNet and PSPNet. Best results are noted with \textbf{bold}. Average Improve. represents average improvement based on coarse mask.}
\label{tab:BIGPerformance}
\end{table*}

\begin{figure}[!t]
  \centering
  \includegraphics[width=0.98\linewidth]{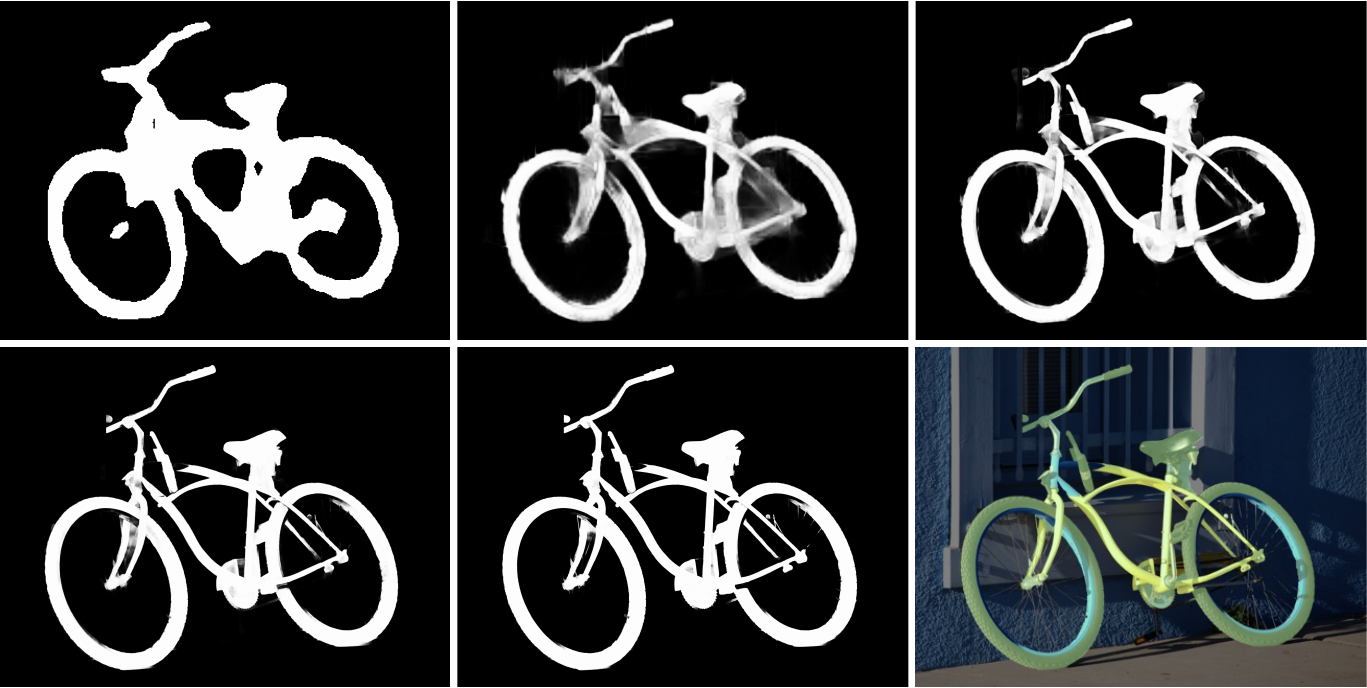}
  \hfill
  \caption{Visualization of refinement steps in our inference strategy. From left to right, top to down: $M_{\text{coarse}}$, refined mask $M^{i}_{\text{refined}}, i\in \{1,2,3,4\}$~(The rescale ratios are 0.125, 0.25, 0.5, and 1.0 here.), and overlay $M^{4}_{\text{refined}}$ on the original image. }
  \label{fig:Refinement}
\end{figure}

\paragraph{Inference Strategy}
\label{sec:Inference}
For the resolution gap between low-resolution in training~(300$\sim$1K) and ultra high-resolution~(2K$\sim$6K) in testing, we propose multi-resolution inference to exploit CRM's continuous $P$ and aligned $F_{cont.}$ fully. The lower part of \cref{fig:GeneralFramework} shows the resolution contrast. Due to the continuous property of CAM, for one image, we can generate outputs of the same target ultra high-resolution $M_{\text{refined}}^{i}$ from multi-resolution input $R^{i}(I_{\text{coarse}}^{i})$.  

In the beginning, inference is around the resolution of training images, and gradually increases input's resolution along the continuous ratio axis $Rs$~(with infinite different rescale ratio) as illustrated in \cref{fig:GeneralFramework}. In particular, we concatenate the original ultra high-resolution image $I$ and the coarse mask $M_{\text{coarse}}$~(initial stage) or refined mask $M_{\text{refined}}^{i-1}$ in previous stage. We rescale it on rescale ratio by $R^{i} \in {Rs}$ to be $I_{\text{coarse}}^{i}$. After refinement, $M_{\text{refined}}^{i}$ is generated and used as $M_{coarse}^{i+1}$ for the next rescale ratio stage. The progressive progress is illustrate as~\cref{eq:test1,eq:test2,eq:test3}:
\begin{equation}
I_{\text{coarse}}^{0} = [I, M_{\text{coarse}}^{0}],
\label{eq:test1}
\end{equation}
\begin{equation}
M_{\text{refined}}^{i}=D_{\phi}\qty(\mathrm{CAM}\qty(E_{\theta}\qty(R^{i}\qty(I_{\text{coarse}}^{i})))),
\label{eq:test2}
\end{equation}
\begin{equation}
I_{\text{coarse}}^{i+1} = [I, M_{\text{refined}}^{i}],
\label{eq:test3}
\end{equation}
where $R^{i}$ is one rescale function of $Rs$, $i$ denotes the refinement stage as the upper right mark. For simplicity, \cref{eq:test2} does not include aggregation. In practice, we select enough $R^{i}$s as required regarding performance or by supporting resource. The relation between performance and the number of $R^{i}$ is illustrated in \cref{fig:cont_sample}. And \cref{fig:Refinement} is an example.

This strategy can also be regarded as a variant of coarse-to-fine operations, where methods~\cite{cheng2020cascadepsp, yang2020meticulous} realize it through cascade in decoder, and method of~\cite{huynh2021progressive} through moving window size in range~(256, 512, 1024, and 2048). They can also use this strategy to shrink the gap. Nevertheless, the relatively heavy cascade-based network and many forward times in inference design hinder their usage. Take CascadePSP~\cite{cheng2020cascadepsp} as example, CascadePSP uses the whole ResNet-50~\cite{he2016deep} as backbone, but CRM use it without conv5$\_$x. Then, the cascade-based decoder in CascadePSP~(three resolution up-samplings and the corresponding computation) is more costly than CRM's CAM and $D_{\phi}$~(a five-layer MLP). Therefore, even with multi-resolution inference, the whole refinement process of CRM can be more than twice as fast as CacadePSP~\cite{cheng2020cascadepsp} in ~\cref{tab:BIGCost}. 


\section{Experiments}
\label{sec:Experiment}
In this section, we evaluate our CRM and compare it with other corresponding state-of-the-art methods on BIG~\cite{cheng2020cascadepsp}, COCO~\cite{lin2014microsoft} and relabeled PASCAL VOC 2012~\cite{everingham2015pascal}. We evaluate the Intersection over Union~(IoU), mean Boundary Accuracy~(mBA)~\cite{cheng2020cascadepsp}, panoptic quality~(PQ)~\cite{kirillov2019panoptic} and average precision~(AP) to measure the ability. Then, we present visualization along with ablation studies to understand the effectiveness of our CRM.

\subsection{Datasets and Methods of Comparison}

For training datasets, we follow the setting of CascadePSP~\cite{cheng2020cascadepsp}. MSRA-10K~\cite{cheng2014global}, DUT-OMRON~\cite{yang2013saliency}, ECSSD~\cite{shi2015hierarchical}, and FSS-1000~\cite{li2020fss} are merged into the training datasets, consisting of 36,572 images with diverse semantic classes~(\textgreater 1,000 classes). For the testing datasets, CascadePSP~\cite{cheng2020cascadepsp} proposes an high-resolution image segmentation dataset, named BIG, for evaluation in ultra high-resolution. The image resolution in BIG ranges from 2K to 6K. To prove that our proposed model is general, we evaluate CRM as the extension of Panoptic Segmentation~\cite{li2021fully} and Entity Segmentation~\cite{qi2021open}. We also evaluate CRM on relabeled PASCAL VOC 2012, which is introduced in~\cite{cheng2020cascadepsp}.

We choose CascadePSP~\cite{cheng2020cascadepsp} as the main comparison method on ultra high-resolution. MGMatting~\cite{mgmatting} is chosen as mask-guided matting method and Segfix~\cite{yuan2020segfix} as a high-resolution segmentation refinement method. PanopticFCN~\cite{li2021fully} and Entity Segmentor~\cite{qi2021open} make benchmark of panoptic and entity segmentation. Our proposed method performs better in terms of precision and speed in almost all experiments, especially on high-resolution images.

\subsection{Implementation Details}
We implement our model with PyTorch~\cite{paszke2019pytorch}, and use ResNet-50~\cite{he2016deep} without conv5$\_$x as our $E_{\theta}$. For training, we use Adam~\cite{kingma2014adam} with $2.25\times 10$$^{-4}$ learning rate. The learning rate is reduced to one-tenth at steps 22,500 and 37,500 in a total of 45,000 steps. The training input concatenates 224 $\times$ 224 patches cropped from the original images and their corresponding perturbed masks. The perturbed masks are randomly perturbed on ground truth with a random IoU threshold between 0.8 and 1.0.

For evaluation, we select 4 rescale ratios from a continuous range to refine in experiments. The total inference time of CRM is still much less than half of CascadePSP~\cite{cheng2020cascadepsp}.

\begin{figure*}[!ht]
	\centering
	\includegraphics[width=0.98\linewidth]{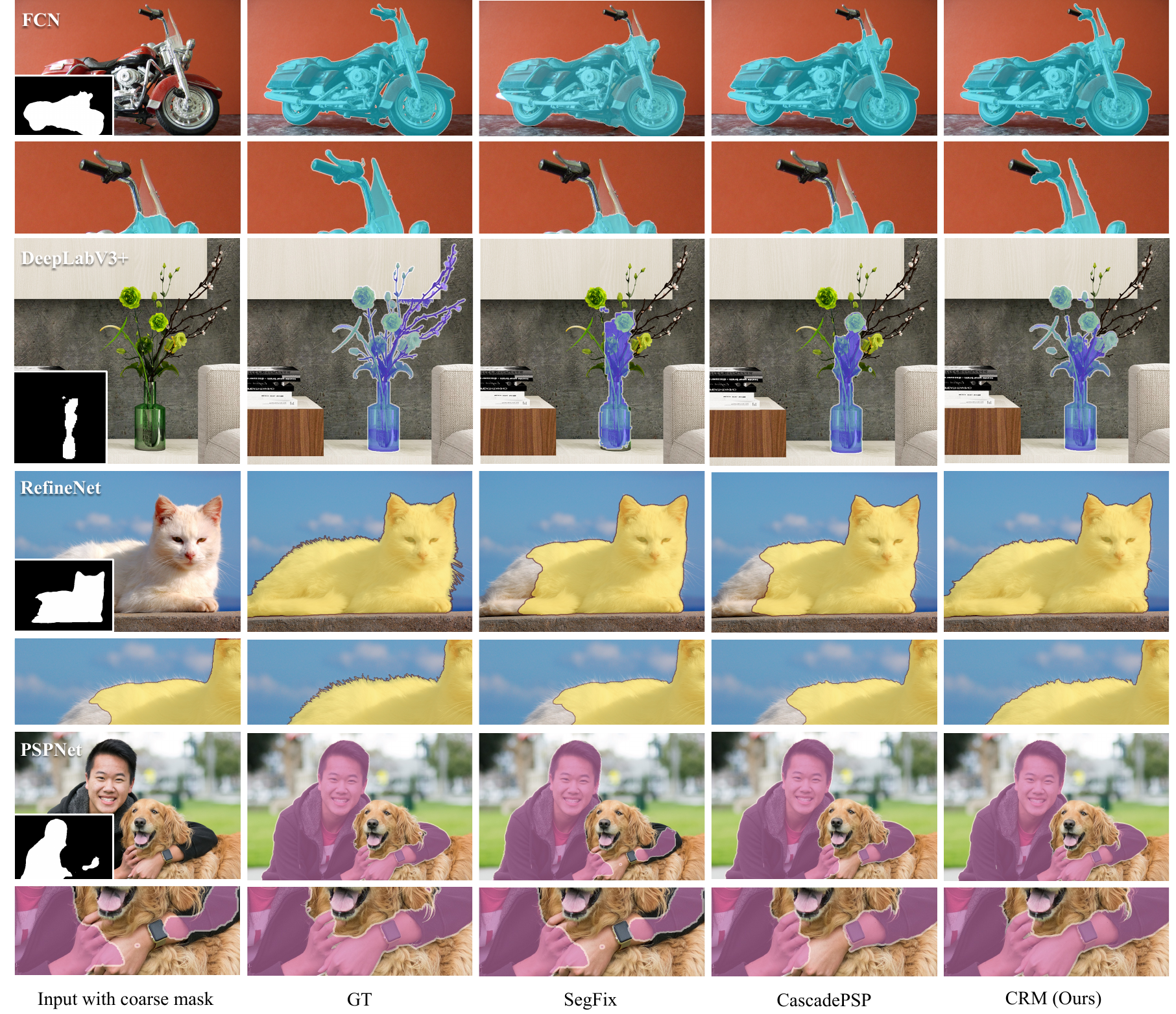}
	\hfill
	\caption{Qualitative comparison between Segfix, CascadePSP and CRM on the coarse mask from FCN, DeepLabV3+, RefineNet and PSPNet. The images are from BIG~(2K $\sim$ 6K). And the black-white mask in bottom left part of first column is the coarse mask.}
	\vspace{-2mm}
	\label{fig:vis}
\end{figure*}

\begin{table}[]
\small
\centering
\begin{tabular}{lccc}
 \toprule
Method~\footnotesize{(IoU/mBA)}   & Time(s) & FLOPs(G) & Params(M) \\ \toprule
CasPSP~\footnotesize{(93.9/75.3)}\cite{cheng2020cascadepsp}  &  620    &   26518    &   67.62   \\ 
CRM~\footnotesize{(94.2/76.1)}&   425  &  2536   &   9.27    \\ 
CRM*~\footnotesize{(93.9/76.3)} &   259  &  1331   &   9.27   \\ \toprule
\end{tabular}
\vspace{-2mm}
\caption{Comparison of total inference time, FLOPs, and the number of parameters on the BIG dataset. CasPSP denotes CascadePSP and selects patches to compute. CRM computes on all pixels. CRM* is a computational-friendly version by just computing the region of interest. Time is recorded on the whole BIG dataset. FLOPs are tested on the same image~(2560*1706).}
\label{tab:BIGCost}
\end{table}

\subsection{Quantitative Results}
In \cref{tab:BIGPerformance} and \cref{tab:BIGCost}, we show comparison among our CRM, CascadePSP~\cite{cheng2020cascadepsp}, Segfix~\cite{yuan2020segfix}, and MGMatting~\cite{mgmatting}. (SegFix and MGMatting perform better on a rescaled image with a downsample ratio 0.5.) They prove that CRM’s performance is better, and it runs faster on high-resolution. 
All segmentation refinement models are trained on low-resolution images and tested on high-resolution images. Segfix and MGMatting's refinement performances are not as good as other methods without a special design for ultra high-resolution images in BIG~\cite{cheng2020cascadepsp}. CascadePSP~\cite{cheng2020cascadepsp} gains more IoU after refinement. Moreover, our CRM produces the highest-quality refinement.  

Besides, the inference time is essential for the ultra high-resolution task. \cref{tab:BIGCost} shows that CRM takes less than half inference time of CascadePSP~\cite{cheng2020cascadepsp} on the whole BIG dataset. FLOPs and parameters are also less. This advantage is due to the simplicity of CRM. 


\begin{table}[]
\small
\centering
\begin{tabular}{lclc}
\toprule
Method                &   PQ  &  Method              &  AP  \\ \toprule
PanopticFCN~\cite{li2021fully}  &  41.0 &  EntitySeg~\cite{qi2021open}  & 38.1 \\ 
PanopticFCN+CRM &  41.8 &  EntitySeg+CRM & 38.9 \\  \toprule
\end{tabular}
\vspace{-2mm}
\caption{The performance after extending PanopticSeg and EntitySeg with our CRM without finetuning.}
\label{tab:PanopticPerformance}
\end{table}

The experiments on panoptic segmentation and entity segmentation are illustrated in \cref{tab:PanopticPerformance}. After adding CRM to \cite{li2021fully} and \cite{qi2021open}, their segmentation performance is enhanced.

We also report our performance on relabeled Pascal VOC 2012 in \cref{tab:relabelPascal}. Compared with CascadePSP~\cite{cheng2020cascadepsp} and Segfix~\cite{yuan2020segfix}, CRM runs better than Segfix~\cite{yuan2020segfix} and is comparable with CascadePSP on IoU, but tends to emphasize more on details. 

These quantitative results show CRM's general effectiveness on ultra high-resolution images as well as low-resolution ones.

\subsection{Qualitative Results}
We show comparison among CascadePSP~\cite{cheng2020cascadepsp}, Segfix~\cite{yuan2020segfix} and our proposed CRM in \cref{fig:vis}. There are more details in our refinement results. It generates matting-style results with only semantic segmentation annotation in training—the matting benefits from continuous alpha-value supervision. Further, the missing part in coarse masks can be reconstructed better through CRM. 

\begin{figure}[t]
  \centering
  \includegraphics[width=0.98\linewidth]{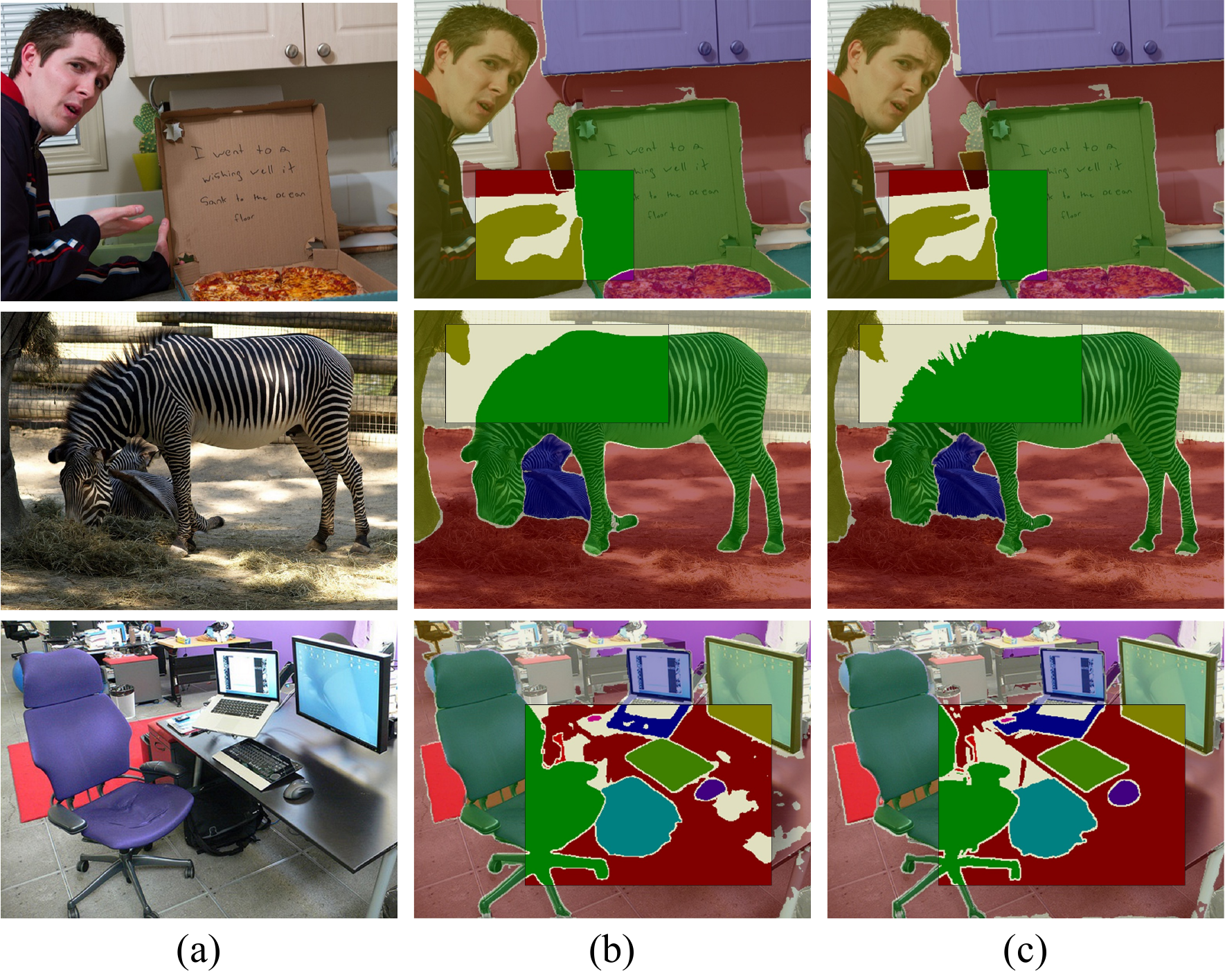}
  \hfill
  \vspace{-1mm}
  \caption{CRM applied in panoptic segmentation. (a) Input image, (b) coarse panoptic segmentation mask, (c) refined mask by our CRM. The images are from COCO.}
  \label{fig:panoptic}
\end{figure}

In addition, we show some visualization of applying CRM into panoptic segmentation in \cref{fig:panoptic}. We can see the mask details and overall segmentation are considerably improved. More results in supplement material further manifest the effectiveness of CRM and the continuous modeling.

\begin{table}[!t]
    \vspace{-4mm}
	\small
	\centering
	\begin{tabular}{lcccc}
		\\\toprule
		IoU/\underline{mBA}  & CM & SF\cite{yuan2020segfix} & CasPSP~\cite{cheng2020cascadepsp} & CRM   \\ \toprule
		FCN-8s~\cite{long2015fully}     & 68.85 & 70.02 & 72.70  &   73.74   \\ 
		& \underline{54.05} & \underline{57.63} & \underline{65.36}  &   \underline{67.17}  \\ \hline
		DeepLab    & 87.13 & 88.03 & 89.01  &   88.33    \\ 
		V3+~\cite{chen2018encoder}        & \underline{61.68} & \underline{66.35} & \underline{72.10}  &   \underline{72.25}  \\ \hline
		RefineNet~\cite{lin2017refinenet} & 86.21 & 86.71 & 87.48  &   87.18 \\ 
		& \underline{62.61} & \underline{66.15} & \underline{71.34}  &   \underline{71.54} \\ \hline
		PSPNet~\cite{zhao2017pyramid}    & 90.92 & 91.98 & 92.86  &  92.52 \\
		& \underline{60.51} & \underline{66.03} & \underline{72.24}  &   \underline{72.48} \\ \toprule
	\end{tabular}
	\vspace{-1mm}
	\caption{Quantitative comparison on relabeled PASCAL VOC 2012. Due to the limited width, CM represent coarse mask, SF represents SegFix, and CasPSP denotes CascadePSP.}
	\label{tab:relabelPascal}
\end{table}

\subsection{Ablation Study}
\vspace{-1mm}
\paragraph{CRM and Inference Resolutions}
CAM and implicit function are the key contributions of our work. The rows in \cref{tab:ExistingofCRM} shows the existence of CRM and implicit function can enhance the performance on every resolution~(the first column means the rescale ratios on $I_{\text{coarse}}^{i}$).  

For the inference strategy, we analyze the columns of \cref{tab:ExistingofCRM}. CRM refines a good general mask at low-resolution ~(IoU mainly increased in low resolution). As the resolution grows, more details are generated, and mBA increases.

\begin{table}[!ht]
    \small
    \renewcommand*{\arraystretch}{0.9}
    \centering
    \begin{tabular}{ccc}
    \toprule
    IoU/mBA  & w/o  CAM\&Impl. & w  CAM\&Impl. \\ \toprule
    0.125    & 92.68/63.70       &   93.07/65.61   \\ 
    0.25     & 93.49/69.23       &   93.88/71.41   \\ 
    0.5      & 93.85/73.43       &   94.15/74.95   \\ 
    1.0      & 93.94/75.42       &   94.18/76.09   \\ \toprule
    \end{tabular}
    \vspace{-2mm}
    \caption{The effect of CRM and inference resolutions. Impl. denotes implicit function.}
    \label{tab:ExistingofCRM}
\end{table}

\begin{table}[!ht]
	\small
	\renewcommand*{\arraystretch}{0.9}
	\centering
	\begin{tabular}{c c c c}
		\toprule
		CAM & Impl. & IoU & mBA  \\ \toprule
		$\times$ & $\times$ &   93.94 & 75.42      \\ 
		$\surd$ & $\times$  &   93.99  &  75.93      \\ 
		$\times$ & $\surd$  &  93.96   &  75.55      \\
		$\surd$ & $\surd$ &  94.18 & 76.09      \\  \toprule
	\end{tabular}
	\vspace{-2mm}
	\caption{The ablation study about CAM and implicit function.}
	\label{tab:ContAlignAndImplicitFunc}
\end{table}

\vspace{-3mm}
\paragraph{CAM and Implicit Function}
The results in \cref{tab:ContAlignAndImplicitFunc} show CAM and implicit functions are all indispensable parts of CRM. Together, they achieve synergy effects.

\begin{figure}[t]
  \centering
  \includegraphics[width=0.98\linewidth]{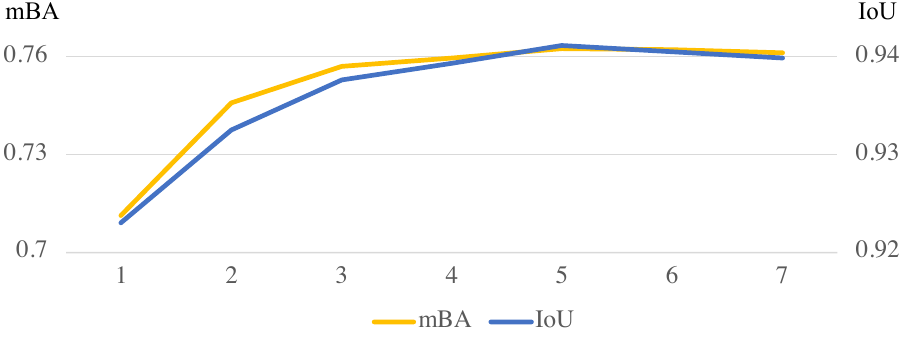}
  \hfill
  \vspace{-1mm}
  \caption{The effect of inference's continuity. The horizontal axis represents the number of uniformly sampled points between 0 and 1. The sampled points are rescale ratios of input.}
  \label{fig:cont_sample}
\end{figure}

\vspace{-3mm}
\paragraph{The effect of inference's continuity}
From \cref{fig:cont_sample}, we can see the performance is growing with the number of sampled rescale ratios between 0 and 1. More numbers mean more continuity in the resolutions of inference, which helps improve performance until convergence. Different from the chosen rescale ratios in ~\cref{fig:Refinement} and \cref{tab:ExistingofCRM}, the final performances are almost the same level as ~\cref{fig:Refinement} and \cref{tab:ExistingofCRM}.

\section{Conclusion}
\vspace{-1mm}
We have proposed CRM to refine segmentation on ultra high-resolution images. 
CRM continuously aligns the feature map with the refinement target, which helps aggregate features for reconstructing details on the high-resolution mask. Besides, our CRM shows its significant generalization potential regarding low-resolution training and ultra high-resolution testing. Experiments show that continuous modeling is promising in terms of performance and speed. 

\noindent \textbf{Limitations} We use the configuration of ``low-resolution training and ultra high-resolution testing'' at present. Using ultra high-resolution images to train and test is still resource-consuming. Addressing this challenging problem will be our future work.

{\small
\bibliographystyle{ieee_fullname}
\bibliography{PaperForReview}
}

\end{document}